\documentclass[]{spie}  %>>> use for US letter paper

\usepackage{amsmath,amsfonts,amssymb}
\usepackage{subcaption}
\usepackage{graphicx}
\usepackage{cite} 
\usepackage{epsfig}
\usepackage{nccmath}
\usepackage{xcolor,colortbl}
\usepackage{tabularx}
\usepackage{relsize}
\usepackage{pifont}
\usepackage{booktabs} 
\usepackage{multirow}
\usepackage{multicol}
\usepackage{adjustbox}
\usepackage{float}
\usepackage{makecell}
\usepackage{tabu}
\usepackage[colorlinks=true, allcolors=blue]{hyperref}
\usepackage[capitalize]{cleveref}
\usepackage{algorithm}
\usepackage{parskip}
\usepackage{svg}
\usepackage{pdfpages}

\title{Organoid Tracker: A SAM2-Powered Platform for Zero-shot Cyst Analysis in Human Kidney Organoid Videos}

\author[a]{Xiaoyu Huang}
\author[b]{Lauren M Maxson}
\author[b]{Trang Nguyen}
\author[b*]{Cheng Jack Song}
\author[a*]{Yuankai Huo}

\affil[a]{Vanderbilt University, Nashville TN 37235, USA}
\affil[b]{University of Alabama at Birmingham, Birmingham, AL 35294, USA}

\authorinfo{*Co-corresponding authors: \\
E-mail: song1c@uab.edu, yuankai.huo@vanderbilt.edu}

\begin{document}
\maketitle

\begin{abstract}

Recent advances in organoid models have revolutionized the study of human kidney disease mechanisms and drug discovery by enabling scalable, cost-effective research without the need for animal sacrifice. Here, we present a kidney organoid platform optimized for efficient screening in polycystic kidney disease (PKD). While these systems generate rich spatial-temporal microscopy video datasets, current manual approaches to analysis remain limited to coarse classifications (e.g., hit vs. non-hit), often missing valuable pixel-level and longitudinal information. To help overcome this bottleneck, we develop Organoid Tracker, a graphical user interface (GUI) platform designed with a modular plugin architecture, which empowers researchers to extract detailed, quantitative metrics without programming expertise. Built on the cutting-edge vision foundation model Segment Anything Model 2 (SAM2), Organoid Tracker enables zero-shot segmentation and automated analysis of spatial-temporal microscopy videos. It quantifies key metrics such as cyst formation rate, growth velocity, and morphological changes, while generating comprehensive reports. By providing an extensible, open-source framework, Organoid Tracker offers a powerful solution for improving and accelerating research in kidney development, PKD modeling, and therapeutic discovery. The platform is publicly available as open-source software at \url{https://github.com/hrlblab/OrganoidTracker}.

\end{abstract}


\section{Introduction}

The advent of human pluripotent stem cell (hPSC) derived organoids has marked a paradigm shift in biomedical research, offering new models that recapitulate key aspects of human organogenesis and pathophysiology~\cite{cable2022organoids}. This new approach provides a crucial bridge between simplistic 2D cell cultures and complex animal models, which often fail to fully mirror human diseases like polycystic kidney disease (PKD)~\cite{liu2022studying}. Kidney organoids, in particular, have emerged as invaluable tools for investigating genetic disorders, enabling the study of cyst formation in a controlled, human-relevant context~\cite{cruz2017organoid}. This potential was recently advanced by the development of scalable platforms capable of efficiently generating thousands of uniform kidney organoids, which robustly model cystogenesis in PKD1 and PKD2 mutant lines. Such high-throughput systems, designed for large-scale therapeutic screening, generate vast amounts of spatial-temporal video data, presenting new and significant analytical challenges~\cite{tran2022scalable}.

Analyzing this dynamic video data remains a significant bottleneck. While the last generation of foundation models, like the Segment Anything Model (SAM), has revolutionized promptable segmentation for static images, an image is merely a "static snapshot of the real world." Organoid development is a temporal process where cysts undergo significant changes in appearance due to motion, deformation, and occlusion, presenting unique challenges that image-based models are not equipped to handle~\cite{ravi2024sam}. Consequently, even with automated bright-field imaging used in large-scale screens, phenotypic scoring is often limited to laborious, categorical outcomes like "positive hit" or "non-specific hit"~\cite{tran2022scalable}.This overlooks the rich quantitative details on cyst initiation and growth dynamics contained within the videos. Furthermore, other automated approaches often require significant computational expertise or are tailored to specific imaging modalities like immunofluorescence or MRI, not the widely used label-free bright-field microscopy used in platforms like that of Tran et al~\cite{park2022deep,monaco2024ai,deininger2023ai}. This creates a critical need for an accessible platform that can perform robust, longitudinal analysis of organoid videos without specialized skills.

\begin{figure}[!h]
    \centering
    \includegraphics[width=0.9\textwidth]{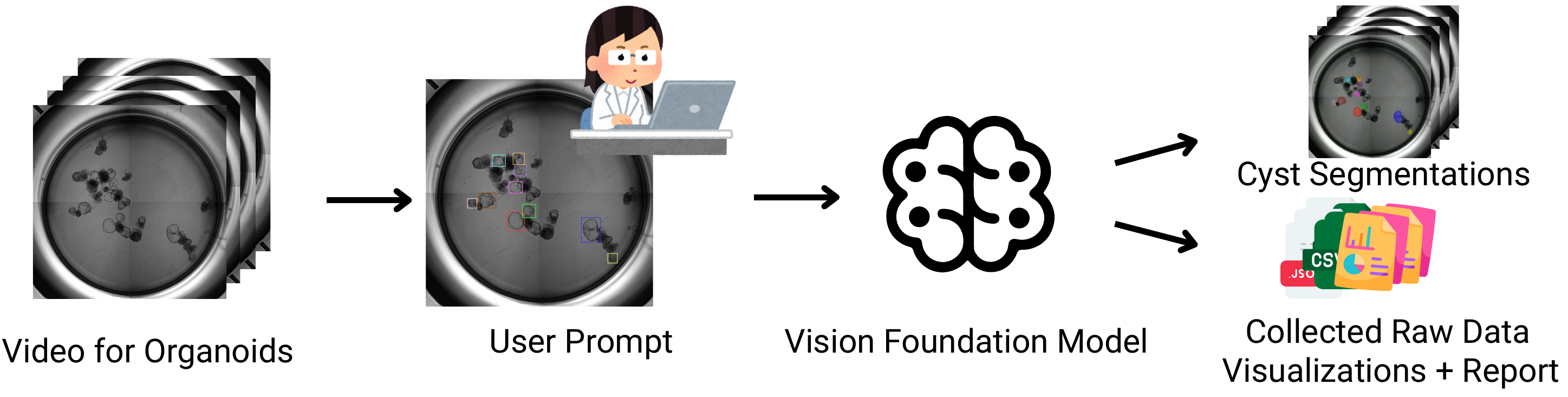}
    \caption{Conceptual Workflow of Organoid Tracker. The workflow begins with user-provided bounding box prompts on a single frame of a time-lapse organoid video. A vision foundation model (SAM2) processes the entire video to generate instance segmentation masks for each cyst over time. From these segmentations, the platform automatically extracts quantitative raw data, producing visualizations and a final analysis report.}
    \label{fig1}
\end{figure}

To address this gap, we developed Organoid Tracker as shown in Fig.~\ref{fig1}, a novel software platform with a graphical user interface (GUI) for the zero-shot analysis of kidney organoid videos. Our platform is built upon the Segment Anything Model 2 (SAM2), a unified foundation model designed for promptable segmentation in both images and videos~\cite{ravi2024sam}. This enables researchers to perform automated segmentation and tracking of cystic structures directly from bright-field microscopy, eliminating the need for fluorescent labels or programming. Beyond an GUI for SAM2 model, the proposed platform is customized for organoid video analyses by employs an innovative inverse temporal tracking strategy, where cysts are first identified in the final, clearest frame and then tracked backward in time. This approach effectively overcomes the difficulty of segmenting nascent cysts as they emerge from dense organoid tissue. Moreover, Organoid Tracker extracts a suite of quantitative metrics, including (1) individual cyst growth trajectories, (2) circularity evolution, and (3) population-level statistics like cyst formation rate and density. These detailed analytics provide an unprecedented view into the dynamics of PKD pathophysiology, enabling a deeper analysis of therapeutic effects beyond simple endpoint classifications~\cite{tran2022scalable}. By offering an open-source, extensible, and user-friendly framework, Organoid Tracker provides a powerful solution to accelerate data analysis in studies of kidney development, disease modeling, and the discovery of novel therapeutics.

\section{Methods}

The overall method for kidney organoid cyst segmentation and analysis is a three-stage process presented in Fig.~\ref{fig1}. First, the user provides input by loading a time-lapse microscopy video and annotating the organoids and cysts on the final frame of the video. Second, the video is processed by a user-selected deep learning model. Finally, the platform generates video segmentations and extracts raw data from these results to produce quantitative analysis and visualizations.

\begin{figure}[!h]
    \centering
    \includegraphics[width=0.9\textwidth]{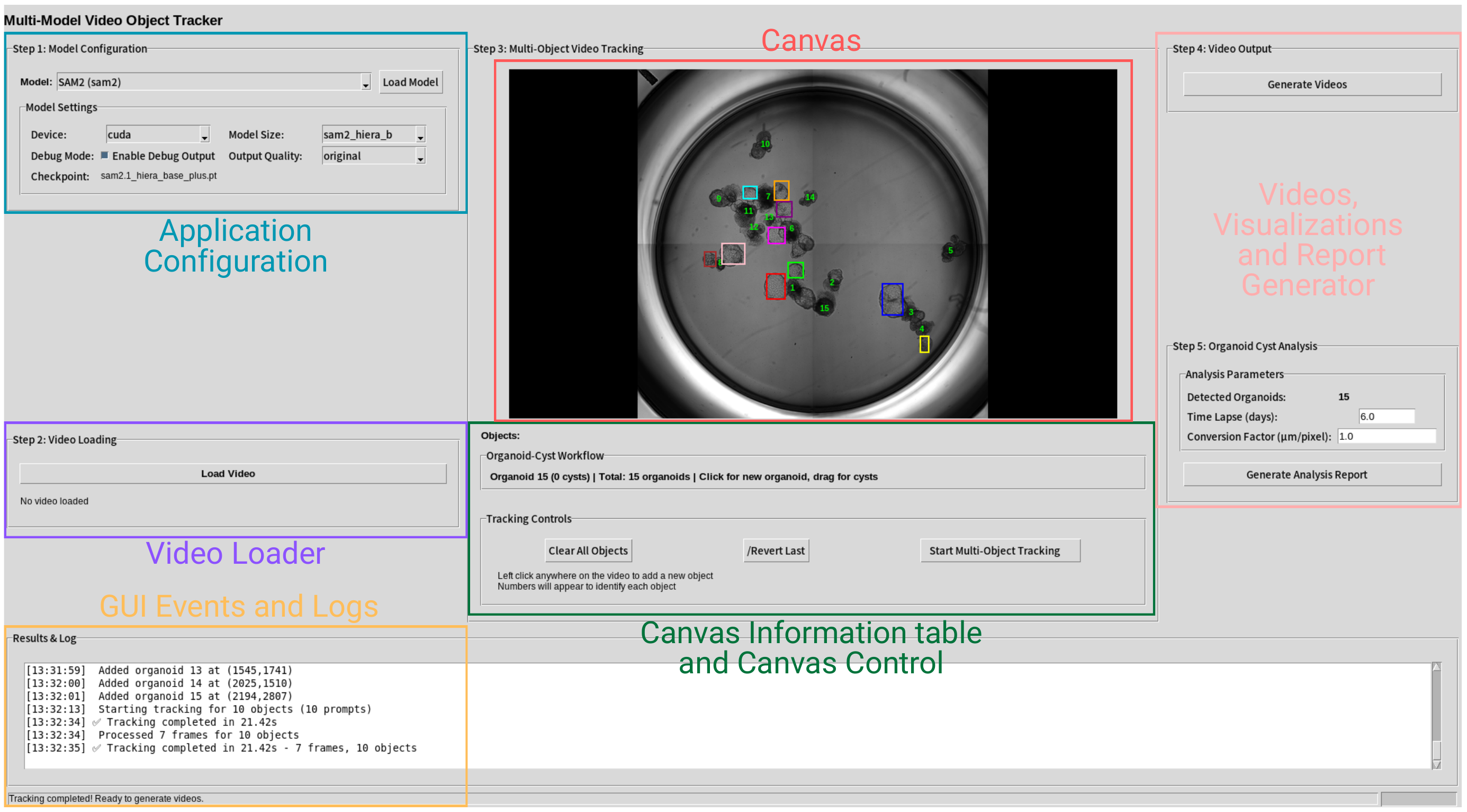}
    \caption{The Organoid Tracker Graphical User Interface (GUI). The user-friendly interface is organized into functional modules: (top-left) application configuration for model selection, (bottom-left) a log for tracking progress, (center) a main canvas for video display and user interaction, and (right) modules for analysis parameter input and report generation.}
    \label{fig2}
\end{figure}

\subsection{GUI User Interface}

To optimize for multi-object tracking, the platform utilizes a bounding box annotation method as shown in Fig.~\ref{fig2}. Initial testing revealed that for low-contrast biomedical imagery, bounding boxes provide a more robust prompt than single points, which can be ambiguous, or class-agnostic automatic segmentation, which often fails to separate clustered cysts. Bounding boxes offer a practical balance between minimal user input and high segmentation accuracy by providing the model with a clear spatial prior for each distinct object of interest.  The user logic is designed to be task-specific for organoid-cyst analysis. The user initiates the process by left-clicking on a parent organoid and then dragging bounding boxes to identify all cysts associated with it. A subsequent left-click designates a new organoid, preparing the system for the next set of cyst annotations. This process establishes a direct association between organoids and their respective cysts, which is crucial for downstream analysis. Annotation is performed on the last frame of the video, as this time point typically provides the clearest view of all cysts that have formed, simplifying their identification.

\subsection{Inverse Temporal Tracking and Data Extraction}

A key innovation of this method is the use of inverse temporal tracking to enhance segmentation accuracy. After the user annotates the final frame, the platform processes the video in reverse chronological order. This backwards-tracking approach is highly effective because it begins with clearly defined cysts, leveraging the temporal memory of models like SAM2 to maintain tracking integrity as the cysts shrink or disappear in earlier frames. This strategy avoids the difficulty of distinguishing newly forming cysts from organoid tissue, a common challenge in forward-pass tracking that yields poor results.

During this process, several key data points are gathered from the segmentation masks at each frame: the total organoid and cyst counts, the association between each cyst and its parent organoid, and the area and circularity of each individual cyst. After the inverse tracking is complete, all temporal data is reversed again to reflect the correct chronological order for final analysis and output generation. To calculate absolute physical measurements, the user manually inputs essential experimental parameters, including the total time-lapse duration and the microscope-specific spatial conversion factor (µm/pixel).

\subsection{Quantitative Spatial-temporal Metrics for Organoid Video Analysis}

The platform generates two primary categories of output: video segmentations for qualitative review and a suite of quantitative data analyses for objective assessment, as previewed in the report generator in Fig.~\ref{fig2}. For visualization, users can generate several types of video outputs, including an overlay of the segmentation masks on the original video, a multi-object mask video, and a side-by-side comparison. To accommodate different needs, such as quick previews of large datasets, the output quality of these videos can be adjusted. From the segmentation data, a suite of quantitative metrics is automatically calculated to analyze the dynamics of cyst formation and development. The following sections detail the core analyses performed.

\subsubsection{Growth Kinetics of Individual Cysts}

To analyze the growth kinetics of each cyst, the platform generates area trajectories that track the morphometric evolution of each cyst's area over the time course (Fig.~\ref{fig3}a). These individual growth curves allow for the direct assessment of growth patterns and rates. This metric provides a direct, high-resolution readout of cyst expansion, the fundamental pathological process driving increased Total Kidney Volume in PKD. This allows for a granular assessment of how interventions impact the core driver of renal function decline. The area (\(A\)) for each cyst at a given time (\(t\)) is calculated from its segmented mask and converted from pixels to physical units (\(\mu m^{2}\)) using the user-provided conversion factor.

\[Area(\mu m^{2})=N_{pixels}\times(ConversionFacter)^{2}\]

\subsubsection{Temporal Analysis of Cyst Morphological Maturation}

To track morphological maturation, the circularity of each cyst is monitored over the experimental time course. This metric provides a quantitative index of cyst maturation, reflecting the transition from an irregular tubular dilation to a structurally stable, well-defined cyst. Monitoring this parameter over time is crucial for understanding how therapeutic interventions affect not only cyst size but also their structural integrity and stability. Circularity is calculated using the standard morphometric formula, where a value of 1 represents a perfect circle and values closer to 0 indicate irregularity (Fig.~\ref{fig3}b).

\[ Circularity(t)=\frac{4\pi\times Area(t)}{Perimeter^{2}(t)} \]

\subsubsection{Population-Level Cystogenesis Metrics}

\begin{figure}[H]
    \centering
    \includegraphics[width=0.85\textwidth]{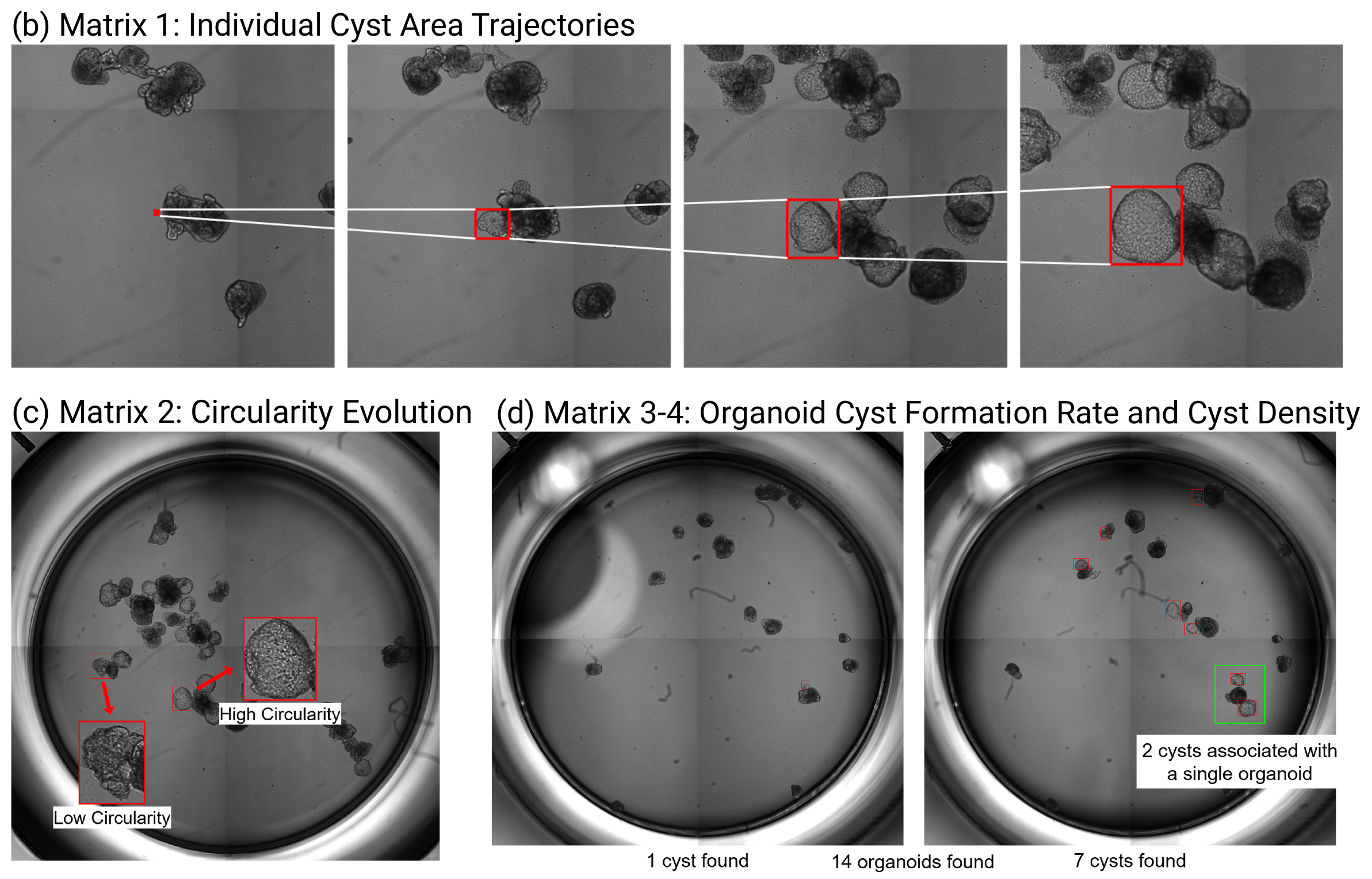}
    \caption{Visual Definition of Key Quantitative Metrics. (a) An individual cyst's cross-sectional area is tracked across multiple time points. (b) Circularity is illustrated by comparing a morphologically irregular cyst (low circularity) with a well-defined, rounded cyst (high circularity). (c) Population metrics are derived by identifying all organoids and their associated cysts at initial and final time points.}
    \label{fig3}
\end{figure}

To assess cyst formation at the population level, the platform calculates two key metrics (Fig.~\ref{fig3}c). The Formation Rate serves as a primary endpoint for identifying interventions that block cyst initiation, a distinct therapeutic goal from slowing the expansion of pre-existing cysts. Furthermore, Cyst Density quantifies the magnitude of the phenotype, allowing for the detection of compounds that may not prevent but significantly reduce the overall cystic burden. The first, Organoid Cyst Formation Rate, measures the percentage of organoids that develop at least one cyst over time. The second metric, Cyst Density, quantifies the intensity of cyst proliferation by calculating the average number of cysts per organoid over time.

\[ Formation Rate(t)=\frac{N_{Organoid With Cysts}(t)}{N_{Total Organoids}}\times 100\% \]

\[ Cyst Density(t) = \frac{N_{TotalCyst}(t)}{N_{TotalOrganoids}} \]

\[ Formation Rate(initial)=\frac{1}{14}\times 100\% , Formation Rate(final)=\frac{6}{14}\times 100\%\]
\[ Cyst Density(initial) = \frac{1}{14} , Cyst Density(final) = \frac{7}{14} \]

\subsubsection{Correlation of Cyst Size, Shape, and Time}

A two-dimensional morphometric analysis is used to visualize the multidimensional relationship between a cyst's shape regularity, its size, and time. This multi-parameter visualization is critical for uncovering the complex relationship between cyst proliferation and structural maturation. By simultaneously plotting size, shape, and time, this analysis can reveal distinct developmental trajectories and identify conditions that may decouple growth from morphological stabilization. This provides deeper mechanistic insights than single-parameter analyses, enabling a more nuanced understanding of how different genetic or chemical perturbations affect the overall cystogenic process. In the scatter plot, each point represents a single cyst at a specific time, with its circularity plotted on the y-axis and time on the x-axis, while the cyst's area is encoded by the color intensity of the point.

\subsubsection{Heatmap of Cyst Growth Heterogeneity}

To visualize population-level heterogeneity, the platform generates a heatmap of area progression across all individual cysts. This analysis directly addresses the critical challenge of heterogeneity in PKD, where cysts exhibit highly variable growth rates and therapeutic responses, likely due to diverse cellular origins and signaling pathway activation. By categorizing cysts into distinct growth phenotypes, the heatmap provides an intuitive, population-level overview that moves beyond simple averages to reveal subpopulations of responders and non-responders. This is essential for identifying broadly effective therapeutics and for dissecting the complex cellular dynamics that govern disease progression. In the heatmap, each row represents a cyst, each column a time point, and color intensity corresponds to the area. Rows are sorted by overall growth rate, facilitating the identification of distinct growth phenotypes like "fast," "medium," and "slow" growing cysts. The growth rate is calculated as the average change in area between consecutive time points, which makes the metric robust to single-frame segmentation anomalies.

\[Overall Growth Rate = \frac{1}{n}\sum_{i=1}^{n}\frac{Area_{i+1}-Area_{i}}{t_{i+1}-t_{i}}\]

\section{Data and Experimental Setting}

From a dataset of time-lapse videos, a representative PKD mutant video and a wild-type control video were selected for a proof-of-concept case study to demonstrate the platform's full analytical workflow. The 7-frame time-lapse bright-field microscopy video used in our case study was generated from kidney organoid cultures. For imaging, organoids were imaged daily for 7 times using an ImageXpress Micro System.

The Organoid Tracker platform was implemented in Python (v3.10) with a graphical user interface built using the native Tkinter library. The core deep learning and computer vision functionalities leverage PyTorch (v2.5.1) and OpenCV (v4.7.0). The platform integrates pre-trained model checkpoints from the Segment Anything Model 2 (SAM2), utilizing both sam2-hiera-base-plus.pth and sam2-hiera-large.pth for segmentation and tracking. All experiments and analyses were conducted on a workstation equipped with an NVIDIA A5000 GPU with 24GB of memory.

\section{Results}

\subsection{Qualitative analysis}

\begin{figure}[H]
    \centering
    \includegraphics[width=1\textwidth]{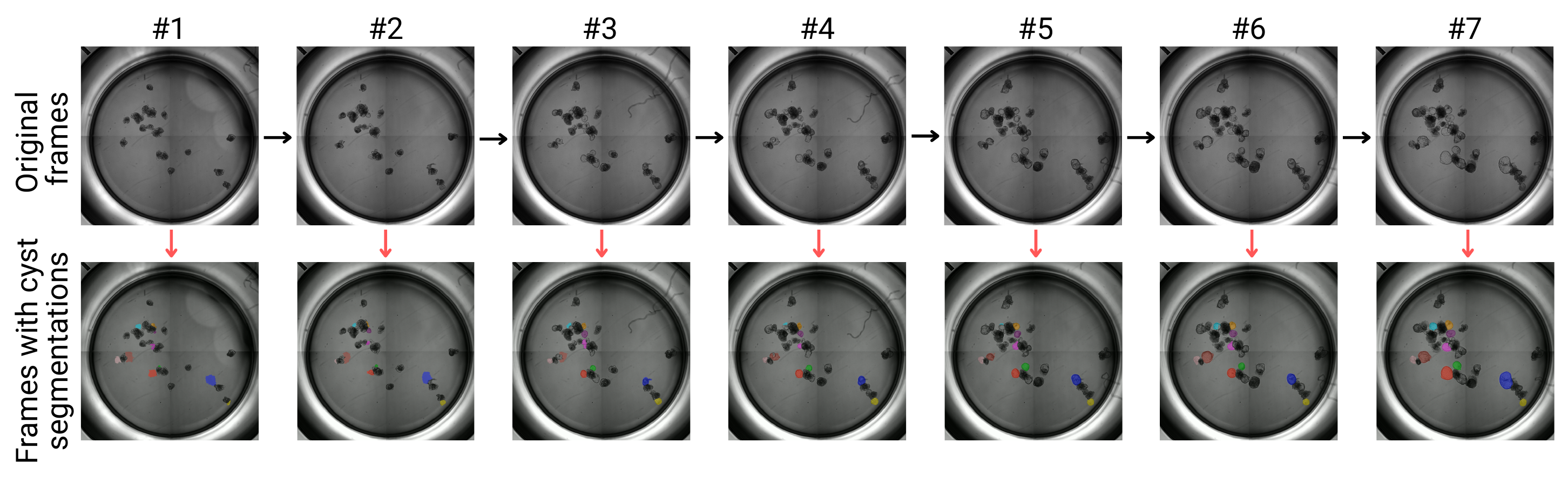}
    \caption{Side-by-side comparison of the original time-lapse video frames (top row) with the corresponding output from Organoid Tracker (bottom row), where automatically generated segmentation masks with unique colors track individual cysts over time.}
    \label{fig4}
\end{figure}

To assess the performance of our platform, we first evaluated the core segmentation and tracking capability of the integrated zero-shot SAM2 model using our inverse temporal tracking method. As illustrated in Figure 4, the platform successfully generated segmentation masks for individual cysts and tracked them over time. The inverse tracking approach, which begins with the clearest view of the cysts in the final chronological frame, yielded segmentation masks with clear and precise boundaries for the initial steps of the reverse-time analysis (corresponding to the later frames of the experiment in Fig.~\ref{fig4}). This demonstrates the model's ability to maintain tracking integrity as cysts decrease in size.

However, we observed mixed results in the final two frames of the reverse-tracking process in Fig.~\ref{fig4}, which correspond to the first two chronological frames of the experiment. In these early frames, where cysts are nascent or have undergone significant positional drift or size changes, the model's performance degraded. This limitation can be attributed to several factors. First, the dataset itself presents challenges, with a low frame count and a large time-lapse between frames, making it difficult for the model's memory mechanism to accurately propagate masks through substantial changes. Second, the zero-shot application of SAM2 encounters a significant domain gap. Foundation models like SAM2 are pre-trained on natural images with strong edge information, which differs substantially from biomedical images that often exhibit "low contrast and weak boundaries."\cite{zhang2024unleashing} This discrepancy leads to performance variability, particularly for challenging targets.

\subsection{Quantitative Analysis}

\begin{figure}[H]
    \centering
    \includegraphics[width=0.95\textwidth]{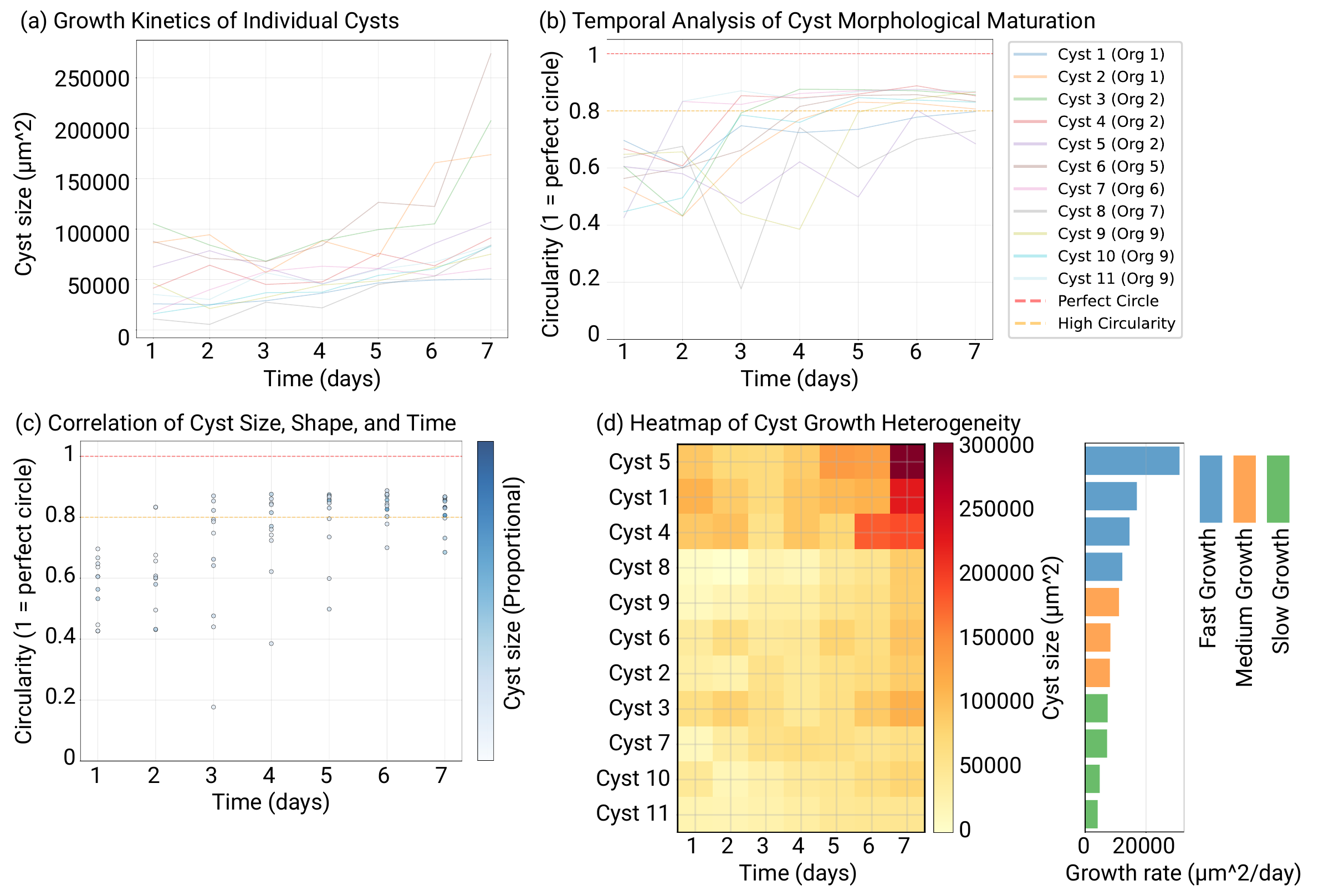}
    \caption{Quantitative Analysis of a Representative PKD Mutant Organoid Video. This figure showcases the analysis suite for a video exhibiting robust cystogenesis. (a) Cyst Area Trajectories: The mean cross-sectional area (\(\mu m^{2}\)) of all cysts, showing progressive growth. (b) Cyst Morphological Maturation: The mean circularity of all cysts shows their evolution towards a more regular shape. (c) Correlation of Cyst Size, Shape, and Time: This scatter plot visualizes the relationship between cyst circularity (y-axis), time (x-axis), and area (color intensity). (d) Heatmap of Cyst Growth Heterogeneity: This visualization provides a population-level overview, with each row representing an individual cyst, sorted by its overall growth rate. The adjacent bar chart quantifies the growth rate for each cyst, enabling the identification of distinct phenotypes.}
    \label{fig5}
\end{figure}

To demonstrate the utility of the platform for downstream analysis, we present a case study showcasing the suite of automatically generated quantitative reports in Fig.~\ref{fig5}5. These reports provide a detailed, multi-faceted view of cyst development, which is critical for objective phenotypic assessment.

$\bullet$ \textbf{Individual Cyst Area and Circularity Trajectories:} The platform tracks the morphometric evolution of individual cysts. The area trajectories (Fig.~\ref{fig5}a) reveal heterogeneous growth kinetics within the population, with most cysts exhibiting a clear upward trend in later time points. Complementing this, the circularity evolution plot (Fig.~\ref{fig5}b) captures morphological maturation. However, it is critical to note that segmentation inaccuracies in the first two chronological frames, where the model struggled to identify nascent cysts with unclear boundaries, introduce significant noise into the initial data points of these trajectories. While many cysts appear to develop into well-defined structures, with circularity values fluctuating but generally remaining high ( > 0.6) in the later, more stable frames, the initial chaotic trends should be interpreted with caution. This is a known failure case consistent with benchmarks where foundation models produce over-segmentation errors during mask propagation\cite{ma2024segment}.

$\bullet$ \textbf{Circularity-Area Correlation:} To further explore the interplay between growth and maturation, the platform generates a multi-parameter correlation plot (Fig.~\ref{fig5}c). This scatter plot visualizes the relationship between a cyst's shape regularity (circularity, y-axis), its size (proportional to point color), and time (x-axis). This analysis successfully visualizes the maturation process, but the data points from the initial two frames are slightly influenced by the aforementioned segmentation errors. For the limited number of cysts and short time course in this case study, clear clusters or tendencies are not immediately apparent, suggesting that longer-term experiments with larger cyst populations would yield more interpretable correlational results.

$\bullet$ \textbf{Heatmap of Cyst Growth:} Finally, the platform provides a population-level overview of growth dynamics using a heatmap (Fig.~\ref{fig5}d). In this visualization, each row represents an individual cyst, each column a time point, and the color intensity corresponds to the cyst's area. Crucially, the rows are sorted by each cyst's overall growth rate, facilitating the clear identification of distinct phenotypes such as "fast," "medium," and "slow" growers. These categories are defined using percentile-based thresholds on the growth rate, ensuring a relative classification adapted to each experiment. The segmentation issues in the initial frames manifest as potentially misleading color blocks at the start of the time course, which may make it difficult to definitively distinguish between cysts with genuinely "slow-growing" phenotypes and those whose growth appears slow due to initial detection failures. This analytical ambiguity highlights why specialized adaptations are often required to improve generalization and ensure reliable quantitative analysis in complex medical imaging tasks.\cite{zhu2024medical}.

$\bullet$ \textbf{Failure Analysis of Population-Level Metrics:} When using the inverse temporal tracking method, the model failed to recognize when a cyst, present in later frames, had not yet formed in earlier frames. Instead of the cyst disappearing from the segmentation map, the model inaccurately propagated the mask to the earliest time points, resulting in erroneously constant values for both Formation Rate and Cyst Density. This failure highlights a challenge for video foundation models in low-frame-rate microscopy: distinguishing true object formation/disappearance from simple morphological changes between sparse frames. This finding suggests that while SAM2 excels at tracking existing objects, supplementary logic or fine-tuning is required for robust event detection like cyst initiation in this domain.

\subsection{Comparative Analysis of PKD Mutant vs. Wild-Type Organoids}

\begin{figure}[H]
    \centering
    \includegraphics[width=1\textwidth]{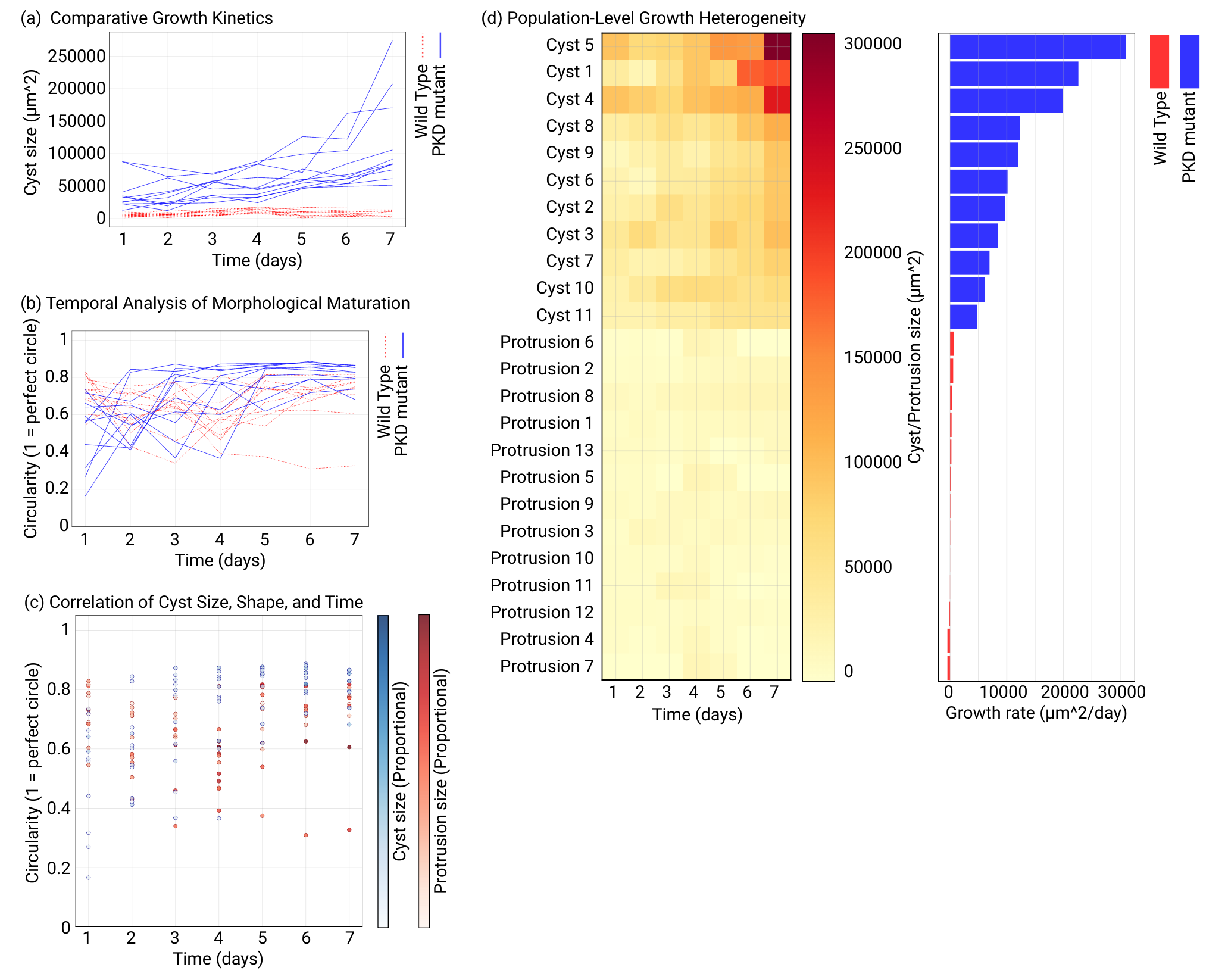}
    \caption{Quantitative Comparison of PKD Mutant and Wild-Type Organoid Phenotypes. Analysis of representative PKD mutant (blue) and wild-type (red) organoids. (a) Mutant cysts show progressive area increase, while wild-type protrusions do not. (b) Mutant cysts achieve higher and more stable circularity than the irregular wild-type structures. (c) In mutants, cyst size correlates with high circularity, a pattern absent in wild-type. (d) The heatmap and growth rate bars confirm distinct population dynamics: sustained growth in mutants versus stasis in wild-type.}
    \label{fig6}
\end{figure}

To demonstrate the platform's capability to distinguish between biological conditions, we performed a comparative analysis between the PKD mutant video and a representative wild-type control. While a visual inspection suggests differences, Organoid Tracker provides a multi-faceted quantitative validation of the phenotypic divergence, transitioning the platform from a computer vision tool to a practical instrument for biological inquiry. The suite of visualizations in Fig.~\ref{fig6} powerfully illustrates these distinctions.

The most striking difference is in the growth kinetics. The trajectories in Fig.~\ref{fig6}a clearly show that while PKD mutant cysts undergo exponential growth over 7 days, the wild-type structures—which we term protrusions rather than cysts due to their distinct morphology—remain stable in size over the period, with trajectories fluctuating around a low baseline. This fundamental difference in growth pattern is the primary indicator of the disease phenotype.

Morphological analysis further separates the two groups. In the temporal analysis of maturation presented in Fig.~\ref{fig6}b, PKD mutant cysts rapidly coalesce into highly regular, spherical shapes, achieving a mean circularity approaching 0.8. The wild-type protrusions, however, exhibit greater intra-group variability and fail to consistently achieve high circularity, suggesting they are transient or structurally unstable entities.

The multidimensional plot in Fig.~\ref{fig6}c reveals a deeper distinction between coordinated and uncoordinated growth. For the PKD mutant, a clear pattern emerges where increasing size (darker color) is strongly correlated with high circularity, indicative of a stable expansion process. Conversely, the wild-type data shows no such correlation; in many cases, the largest protrusions are associated with low circularity, suggesting irregular, non-pathological bloating rather than true cystic growth.

Finally, the population-level overview provided by the heatmap and growth rate chart provides an unambiguous summary of the experiment. The heatmap in Fig.~\ref{fig6}d starkly visualizes the temporal increase in area for the mutant cysts (darkening colors) versus the static nature of the wild-type protrusions. This is quantitatively confirmed by the adjacent bar chart, which shows large, positive growth rates for every mutant cyst, while the growth rates for all wild-type protrusions are centered around zero, with some even showing negative values (shrinkage). This direct comparison demonstrates Organoid Tracker's capability to automatically quantify and visualize distinct, disease-relevant phenotypes, confirming its potential as a powerful tool for objective, high-throughput screening.

\section{Conclusion}

This study introduces Organoid Tracker, a comprehensive platform for the zero-shot segmentation and quantitative analysis of cyst dynamics in kidney organoid videos. By leveraging the SAM2 foundation model with an innovative inverse temporal tracking method, our tool successfully automates the extraction of key morphometric data from bright-field microscopy. Our case study demonstrates the platform's utility, though a performance gap remains in accurately segmenting nascent cysts in early time-lapse frames due to the domain gap. Nevertheless, Organoid Tracker provides a powerful, accessible solution for high-throughput screening, offering valuable insights to accelerate PKD modeling and therapeutic discovery.

% \section{New or breakthrough work to be presented}
% This work presents the first accessible, GUI-based platform that leverages the SAM2 foundation model for zero-shot segmentation and quantitative analysis of cystogenesis in kidney organoid videos, bridging the gap between advanced AI and high-throughput biological research.

\textbf{Acknowledgments} % equivalent to \section*{ACKNOWLEDGMENTS}       

Work in CJS’s laboratory was supported by a grant from STATION 41. Work in CJS was supported by the Polycystic Kidney Disease Research Resource Consortium (PKD RRC) Sprint Challenge Grant, funded by NIDDK Grant U24DK126110. This research was also supported by NIH R01DK135597 (Huo), DoD HT9425-23-1-0003 (HCY), and KPMP Glue Grant. This work was also supported by Vanderbilt Seed Success Grant, Vanderbilt Discovery Grant, and VISE Seed Grant. This project was supported by The Leona M. and Harry B. Helmsley Charitable Trust grant G-1903-03793 and G-2103-05128. This research was also supported by NIH grants R01EB033385, R01DK132338, REB017230, R01MH125931, and NSF 2040462. We extend gratitude to NVIDIA for their support by means of the NVIDIA hardware grant. This work was also supported by NSF NAIRR Pilot Award NAIRR240055.

\textbf{Declaration of interests}

CJS receives research support from Amgen under a sponsored research agreement with UAB. The remaining authors declare no competing interests.

\clearpage

\bibliographystyle{spiebib} 
\bibliography{report.bib}

\end{document}